%% file: aaai2026.tex
\documentclass[letterpaper]{article} 
\usepackage{aaai2026}  

\usepackage{times}  
\usepackage{helvet}  
\usepackage{courier}  
\usepackage[hyphens]{url}  
\usepackage{graphicx} 
\usepackage{natbib}  
\usepackage{caption} 
\usepackage{algorithm}
\usepackage{algorithmic}

\usepackage{newfloat}
\usepackage{listings}

\usepackage{amsmath}
\usepackage{amsfonts}
\usepackage{amssymb} 

\usepackage{multirow}
\usepackage{booktabs}

\DeclareCaptionStyle{ruled}{labelfont=normalfont,labelsep=colon,strut=off} 
\floatstyle{ruled}
\newfloat{listing}{tb}{lst}{}
\floatname{listing}{Listing}
\title{AutoTIR: Autonomous Tools Integrated Reasoning via Reinforcement Learning}
\nocopyright 
\author{
    Yifan Wei\textsuperscript{\rm 1,2}, 
    Xiaoyan Yu\textsuperscript{\rm 3},
    Yixuan Weng\textsuperscript{\rm 4}, Tengfei Pan\textsuperscript{\rm 2},
    Angsheng Li\textsuperscript{\rm 1}\footnotemark[2], 
    Li Du\textsuperscript{\rm 2}\footnotemark[2]
}
\affiliations{
    \textsuperscript{\rm 1}Beihang University
    \textsuperscript{\rm 2}BAAI
    \textsuperscript{\rm 3}Beijing Institute of Technology
    \textsuperscript{\rm 4}Westlake University\\

    weiyifan@buaa.edu.cn, angsheng@buaa.edu.cn, 
    duli@baai.ac.cn

}

\usepackage{bibentry}

\begin{document}

\maketitle

\renewcommand{\thefootnote}{\fnsymbol{footnote}}
\footnotetext[2]{Corresponding Authors.} 

\begin{abstract}
Large Language Models (LLMs), when enhanced through reasoning-oriented post-training, evolve into powerful Large Reasoning Models (LRMs).  
Tool-Integrated Reasoning (TIR) further extends their capabilities by incorporating external tools, but existing methods often rely on rigid, predefined tool-use patterns that risk degrading core language competence.  
Inspired by the human ability to adaptively select tools, we introduce \textbf{AutoTIR}, a reinforcement learning framework that enables LLMs to autonomously decide \textit{whether} and \textit{which} tool to invoke during the reasoning process, rather than following static tool-use strategies.
AutoTIR leverages a hybrid reward mechanism that jointly optimizes for task-specific answer correctness, structured output adherence, and penalization of incorrect tool usage, thereby encouraging both precise reasoning and efficient tool integration.
Extensive evaluations across diverse knowledge-intensive, mathematical, and general language modeling tasks demonstrate that AutoTIR achieves superior overall performance, significantly outperforming baselines and exhibits superior generalization in tool-use behavior. These results highlight the promise of reinforcement learning in building truly generalizable and scalable TIR capabilities in LLMs. The code and data are available at 
https://github.com/weiyifan1023/AutoTIR.

\end{abstract}


\input{text/1-Introduction}
\input{text/6-RelateWork}

\input{text/4-Method}

\input{text/5-Experiments}

\input{text/7-Conclusion}

\bibliography{aaai2026}


\appendix
\input{text/8-Appendix}
\end{document}

%% file: text/1-Introduction.tex
\section{Introduction}
\label{sec: Introduction}

Large Language Models (LLMs) have shown remarkable progress in language understanding~\cite{du2023shortcut}, reasoning~\cite{havrilla2024glore}, and instruction following~\cite{adlakha2024evaluating}, achieving strong performance across diverse natural language processing (NLP) tasks~\cite{minaee2024large}. 
Recent developments in reasoning-oriented post-training, such as reinforcement learning~\cite{chu2025sft}, have further enhanced LLMs' multi-step reasoning capabilities, leading to the rise of Large Reasoning Models (LRMs) that generalize well to increasingly complex problem domains~\cite{openai-o1, guo2025deepseek}.

To extend the reasoning abilities of LLMs beyond language modeling alone, recent research has explored the use of external tools, such as retrieval systems~\cite{song2025r1} and code interpreters~\cite{hosain2025xolver,mai2025agent}, to grant models access to real-time external knowledge and precise numerical computation, two critical shortcomings of pure-text-based LLMs.  
This emerging paradigm, known as Tool-Integrated Reasoning (TIR), has proven effective across a wide range of tasks, enabling LLMs to handle more complex queries, follow longer reasoning chains, and achieve higher task-specific accuracy~\cite{gou2023tora,ma2024sciagent,qin2024tool,zhou2025mem1,lu2025octotools}.
For instance, ReSearch~\cite{chen2025learning} and Search-R1~\cite{jin2025search} augment LRMs with retrieval tools, leading to substantial gains in multi-hop question answering. 
Similarly, MathSensei~\cite{das2024mathsensei} and MathCoder~\cite{wang2024mathcoder} employ code-based execution to improve mathematical reasoning. 
These works demonstrate that invoking external tools can significantly enhance the reasoning abilities of LLMs in domains where factual precision or symbolic manipulation is critical.

\begin{figure}[t]
   \begin{center}
   \includegraphics[width=1\linewidth]{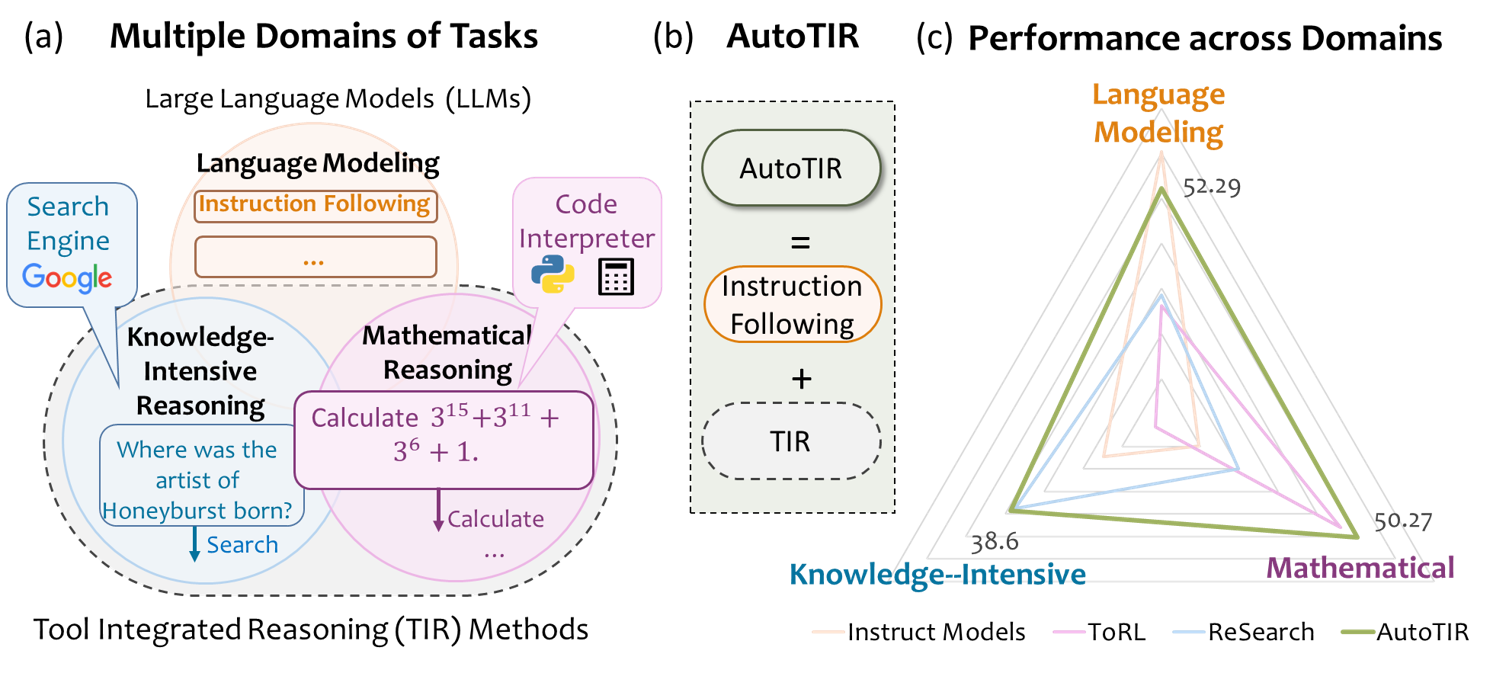}
   \end{center}
   \caption{AutoTIR balances the tool-integrated reasoning with instruction following ability.}
   \label{fig:example}
\end{figure}

However, TIR-based methods often introduce a new set of challenges. 
Many current approaches rely on fixed tool-use patterns, hard-coded templates, or supervised traces, limiting the model’s flexibility across tasks and tools~\cite{yao2023react, wei2023menatqa,  schick2023toolformer, yang2023gpt4tools, li2024chain, feng2025retool, feng2025mt}. 
More critically, learning tool-use strategy is always accompanied by degradation in core language understanding and instruction-following capabilities~\cite{li2025thinking,fu2025scaling}, compromising what makes LLMs broadly useful in the first place.
We argue that these challenges stem from a lack of decision ability to determine whether external tool(s) should be invoked, and which tool should be called.

Inspired by this observation, we introduce AutoTIR, a reinforcement learning framework designed for Autonomous Tools Integrated Reasoning. 
AutoTIR enables the model to autonomously decide both whether to invoke an external tool and which tool to use during the reasoning process. As illustrated in Figure~\ref{fig:example}~(a) and (b), AutoTIR seeks to combine the strengths of LLMs and TIR-enhanced LRMs and avoid their shortcomings \cite{minaee2024large}. 
How to balance the trade-off between language modeling and tool integration comprises the main challenge in learning such a strategy. 
In this paper, we explore addressing this problem using a pure data-driven paradigm by reinforcement learning. 
Specifically, we propose an advantage-based reward system, which i) employs an action reward to supervise the LLM to make correct tool invocation action: on datasets where TIR show significant advantage compared to vanilla instruction model, the model is rewarded to make appropriate tool invocation and punished for redundant tool invocation; while on general domain instructions, the LLM is encouraged to make free exploration about whether tool is necessary and learn from experience; ii) employs output reward to promote model makes responses better than reference model based on the tool invocation action.

We extensively evaluate AutoTIR across a broad range of challenging tasks, encompassing knowledge-intensive, mathematical, and general reasoning domains. As shown in Figure~\ref{fig:example}~(c) and detailed analysis in the Experimental section, AutoTIR achieve consistent performance improvements compared to baseline methods, primarily by learning an autonomous and generalizable tool invocation strategy that effectively balances the tool-integrated reasoning process with core language modeling and instruction following. 
Analysis of tool usage metrics further confirms AutoTIR's capacity for efficient, context-aware tool integration, minimizing superfluous invocations while maximizing successful outcomes, 
and such adaptive tool-use behavior demonstrates a scalable behavior along with its inherent ability to generalize across diverse task demands. These findings underscore the potential of reinforcement learning as a foundation for building generalizable and scalable reasoning in LLMs, by equipping them with the crucial ability to make autonomous, precise tool selection decisions that dynamically respond to problem complexity.

\begin{figure*}[ht]
   \begin{center}
   \includegraphics[width=1\linewidth]{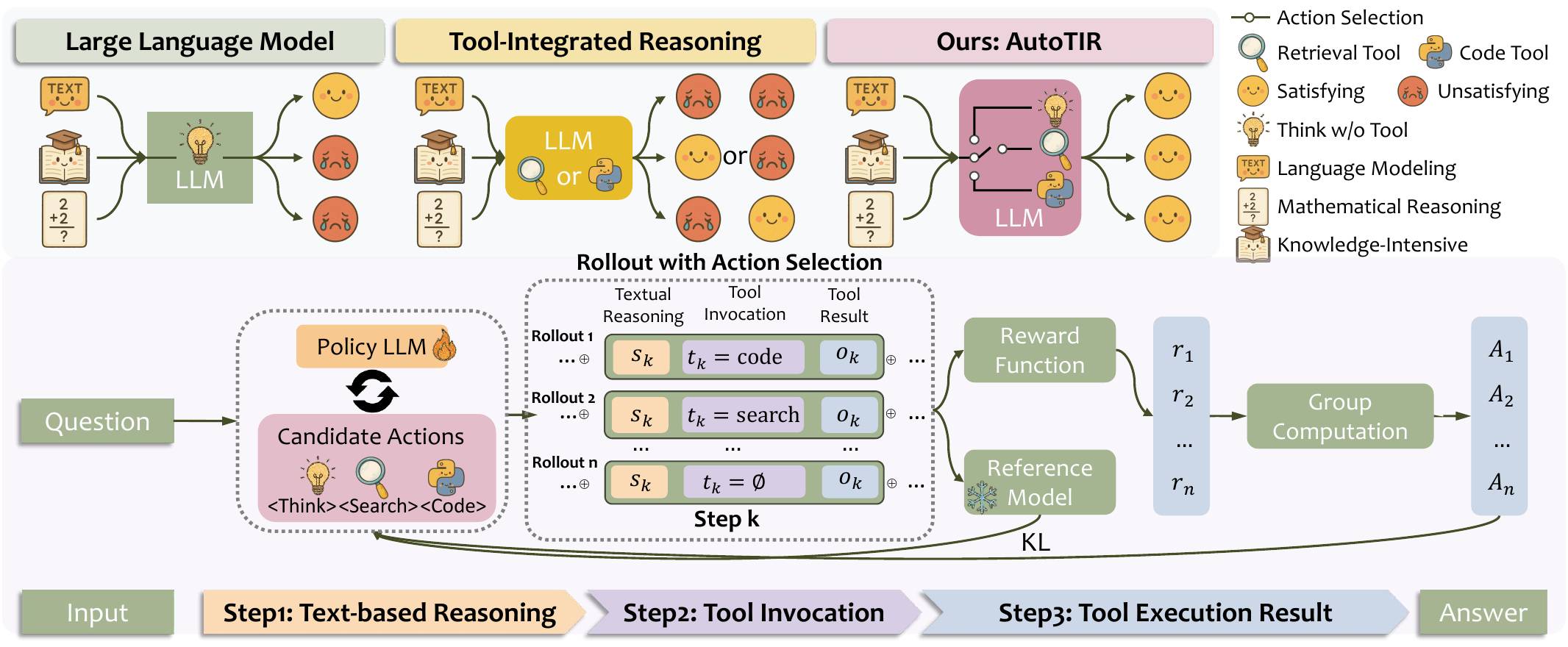}
   \end{center}
   \caption{Overall framework of AutoTIR. 
       \textbf{Top}: Comparison between AutoTIR and existing paradigms (fixed reasoning strategy vs.\ autonomous decision). 
       \textbf{Bottom}: GRPO training pipeline that incorporates multiple reasoning actions.
   }
   \label{fig:framework}
\end{figure*}

%% file: text/6-RelateWork.tex
\section{Related Work}
\label{sec: RelatedWork}

\paragraph{Reinforcement Learning with Verifiable Rewards.}
Reinforcement Learning with Verifiable Rewards (RLVR) offers a promising approach to improve the reasoning of the language model through RL techniques \cite{zelikman2022star,hoffman2023training}. It has proven especially effective for tasks with clear correctness criteria such as solving mathematical problems and code generation.
Rather than relying on learned reward models \cite{ouyang2022training,yeuncertainty}, RLVR employs rule-based verification functions \cite{kazemnejadvineppo, xu2024dpo, wei2025structural, gehring2025rlef}, such as exact answer matching, to generate reward signals. 
This avoids the complexity and potential instability associated with learned reward modeling.
The approach has led to significant progress, exemplified by models such as DeepSeek-R1 \citep{guo2025deepseek}, OpenAI-o1 \citep{openai-o1}, and QwQ \citep{team2024qwq}, which achieve SOTA performance in reasoning tasks.
Furthermore, the development of robust policy optimization algorithms, such as PPO \citep{schulman2017proximal}, DPO \citep{rafailov2023direct}, SimPO \citep{meng2024simpo}, and GRPO \citep{shao2024deepseekmath}, has played a key role in RLVR’s success.

\paragraph{Tool-Integrated Reasoning.}
Tool Integrated Reasoning (TIR) has emerged as a promising paradigm for enhancing the capabilities of LLMs by enabling them to interact with external tools~\cite{li2023api, qin2024toolllm, ye2025tooleyes,liu2025toolace}. Early TIR approaches primarily rely on in-context learning or SFT-based methods~\cite{yao2023react, wei2023menatqa,  schick2023toolformer}.
However, these methods often show limited generalization to unseen tasks or tool configurations.
To address this issue, recent work \citep{jin2025search,song2025r1,li2025torl,feng2025mt,wang2025otc,huang2025reinforced,chen2025learning,dong2025tool} explores training LLMs to acquire a strategy of appropriately utilizing tools using RLVR. Empirical results show that, based on the rule-based rewards, RL enables models to learn how to invoke tools through exploration, often resulting in more robust and adaptive tool-use strategies compared to in-context learning or SFT-based methods.
Despite these advances, most existing efforts are restricted to single-tool settings, where models can only interact with a fixed tool (e.g., code interpreter) in a fixed scenario (e.g., mathematical reasoning). 
Research indicates that the performance improvement upon corresponding benchmarks is always at the cost of instruction following ability and generality~\cite{li2025thinking,fu2025scaling}.  Maintaining the generalization ability of the base model while scaling to multi-tool usage remains a challenging problem.

%% file: text/4-Method.tex
\section{Methodology}
\label{sec: Methodology}

In contrast to prior work that relies on predefined tool(s) for specific domains, our objective is to empower models with the autonomous decision-making capability regarding whether and which tool to invoke across diverse tasks, thereby increasing problem complexity.
We explore addressing this issue using pure reinforcement learning. We begin by formalizing the tool-involved reasoning process, followed by a detailed introduction to the AutoTIR framework.

\subsection{Problem Formalization}
Given a question $Q$ and an environment $E$ that provides access to a set of tools, 
formally, the tool-integrated reasoning process $\tau$ could be defined as:
\begin{equation}
    \tau = \mathcal{A}_1 \oplus, \cdots, \oplus \mathcal{A}_N
\end{equation}
where $\mathcal{A}_k = \langle s_k, t_k, o_k \rangle$ is an intermediate reasoning step, with $s_k$, $t_k$, and $o_k$ denoting a natural language reasoning step, a tool invocation content, and the corresponding execution result of  $t_k$, respectively. Note that, in our formalization, if the model decides that at step $k$, no tool should be invoked, then $t_k=\emptyset$, and consequently $o_k=\emptyset$. Hence, for a question that could be solved using pure natural language, $\forall k\in [1, N], t_k=\emptyset, o_k=\emptyset$.    
The iterative generation process proceeds as follows:
\begin{equation}
\begin{aligned}
& \left(s_k, t_k\right) = M\left(Q \oplus \tau_{k-1}\right) \\
& o_k = E\left(t_k\right) \\
& \tau_k = \tau_{k-1} \oplus  \mathcal{A}_k
\end{aligned}
\end{equation}
This cycle continues until the model $M$ produces a final answer or reaches the maximum context length.

Without loss of generality, we set the available external tools as \textbf{search engine} to retrieve relevant information from local databases, and \textbf{code interpreter} for executing model-generated code snippets in a sandbox environment, returning either the execution results or error messages. Note that more tools could also be accommodated under such problem formalization.

\subsection{Reward Design}
From the above problem formalization, obtaining a correct result relies on two key aspects: 
i) Taking appropriate actions to decide whether a tool should be called, and which specific tool should be utilized at each step.
ii) Generating an accurate final answer that effectively integrates the information gained from tool executions (or pure language reasoning).
To achieve this, we design an advantage-based reward system that incorporates two types of rewards:
i) an \textbf{action reward} to drive the model to make correct tool invocation decisions and learn an optimal tool-use strategy, 
and ii) an  \textbf{output reward} to drive the model to obtain accurate final answers based on its reasoning process.
The total reward is the weighted sum of action reward and output reward:

\begin{equation}
    r=0.1\times r_\text{act} + 0.9 \times r_\text{out}
\end{equation}

\paragraph{Action Reward.}
How to balance the trade-off between language modeling and tool integration comprises the main challenge in learning an appropriate strategy.
To address this, we design an advantage-based action reward mechanism to guide model to make correct tool invocation actions.
In tasks where tool-integrated reasoning provides clear advantages, the model is rewarded for appropriate tool invocation: 
\begin{itemize}
    \item On complex knowledge-intensive tasks, the model is rewarded to utilize the searching engine tool;
    \item On mathematical problems with massive numerical calculation, the model is rewarded to invoke the code interpreter; 
    \item On open domain instances where the benefits of tool invocation are less pronounced (e.g., general instruction-following or simple language-only reasoning), we do not restrict the tool invocation action, the model is encouraged to explore tool utilization to learn when invocation is unnecessary by experience. This promotes efficiency and preserves core language capabilities.
    \item To discourage overuse and misuse of tools, we introduce a specific penalty term $r_\text{penalty}<0$, for incorrect tool selections: on math problems, once searching engine is invoked, and once if the code interpreter is invoked on the knowledge-intensive instances, then a penalty is applied, even if the final answer is correct.

\end{itemize}

Formally, the action reward is defined as:
\begin{equation}
  r_\text{act} = 
  \begin{cases}
    1, & \text{For correct tool invocation}\\
    r_\text{penalty}, & \text{For wrong tool usage} \\
    1, & \text{For open domain instances} \\
  \end{cases}
\end{equation}

\paragraph{Output Reward.}
This reward incentivizes the model to produce accurate final results based on tool invocation actions and subsequent reasoning. To this end, in the output reward $r_\text{out}$, we require the model's output to conform to a specific format indicated by the \verb|\boxed{}| token, and use task-specific evaluation functions to measure the correctness of outputs across different types of tasks. Formally:
\begin{equation}
  r_\text{out} = 
  \begin{cases}
    \text{max}[0.1,\ f_\text{eva}(\text{pred},\ \text{gt})], & \text{If output correctly} \\
     & \quad \quad \quad \ \ \text{formatted} \\
    0, & \text{Else} \\
  \end{cases}
\end{equation}
where $f_\text{eva}(\cdot)$ is the task-specific evaluation function, $\text{pred}$ is the extracted model generated answer, $\text{gt}$ is the ground truth answer.
For question-answering (QA) tasks (which usually appear in the knowledge-intensive reasoning domain), correctness is assessed via a rule-based F1 score, i.e., $f_\text{eva}=\text{F1}(\text{pred},\ \text{gt})$. 
For mathematical reasoning samples, a binary 0/1 reward is assigned for wrong and correct answers, respectively. 
In instruction-following (IF) tasks from general domains, we determine full adherence to the given instructions through an IF score derived from rule matching, granting a reward of 1 for satisfied instructions and 0 otherwise, formally, $f_\text{eva}=\text{IFScore}(\text{pred},\ \text{gt})$.

\subsection{Inference with Multiple Tools.}
We structure a system prompt to explicitly guide the model to perform a predefined reasoning format during the training stage (see Appendix for the full prompt).
In detail, the rollout process proceeds iteratively through three stages: text-based reasoning, tool invocations, and tool execution results, as illustrated in Figure~\ref{fig:framework}. 
Specifically, at the $k$-th reasoning step, the model begins with a \texttt{<think>} phase. If this reasoning leads to a decision to use an external tool, the generation continues with a \texttt{<code>} or \texttt{<search>} tag, corresponding to the code interpreter and the search engine tool, respectively. If the tool tag is \texttt{<code>}, a code execution tool is called; if it's \texttt{<search>}, a search engine tool is invoked. The resulting output is then inserted between \texttt{<result>} and \texttt{</result>} tags and fed back into the rollout for continued generation. This iterative process continues until the model obtains a final answer or reaches the maximum context length.

Note that, AutoTIR can autonomously decide to bypass tool calls when it determines external assistance is not required. In such instances, the model's \texttt{<think>} phase directly leads to the generation of the final answer within \verb|\boxed{}|, entirely skipping the tool invocation stages marked by \texttt{<code>} or \texttt{<search>} . To prevent training bias from environment feedback, execution results from tool invocation are masked during loss computation and do not contribute to gradient updates.

\subsection{RL Learning Algorithm}

We employ Group Relative Policy Optimization (GRPO) \citep{shao2024deepseekmath} as the RL learning algorithm, which estimates the baseline using a group of rollouts.
Given a reference policy $\pi_{{\text{ref}}}$ and an existing policy $\pi_{\theta_{\text{old}}}$, base on $G$ rollouts $\tau = \{y_i\}_{i=1}^{G} \sim \pi_{\theta_{\text{old}}}(\cdot|x)$ for each input $x \sim \mathcal{D}$, the objective of GRPO is to optimize the policy $\pi_{\theta}$ by maximizing the following objective:
\begin{equation}
\begin{aligned}
J_{\mathrm{GRPO}}(\theta) 
&= \mathbb{E}_{x \sim \mathcal{D},\, \{y_i\}_{i=1}^G \sim \pi_{\theta_{\mathrm{old}}}(\cdot \mid x)} \\
&\Biggl[
  \frac{1}{G} \sum_{i=1}^G
  \min\!\Bigl(
    \frac{\pi_{\theta}(y_i \mid x)}{\pi_{\theta_{\mathrm{old}}}(y_i \mid x)}\,A_i,\, \\
    &\mathrm{clip}\!\Bigl(
      \frac{\pi_{\theta}(y_i \mid x)}{\pi_{\theta_{\mathrm{old}}}(y_i \mid x)},
      1-\varepsilon,\,
      1+\varepsilon
    \Bigr)
    A_i
  \Bigr)  \\
  &-\,\beta\,D_{\mathrm{KL}}\bigl(\pi_{\theta}\,\big\|\,\pi_{\mathrm{ref}}\bigr)
\Biggr],
\end{aligned}
\label{eq1}
\end{equation}
where $A_{i} = \left(r_i - \text{mean}(\{r_j\}_{j=1}^{G})\right) / \text{std}(\{r_j\}_{j=1}^{G})$ is the normalized \emph{advantage} of the $i$-th rollout in current group compared to the reference model. Thus, by choosing the reference model as an instruction model, AutoTIR could optimizes the policy model to improve the \emph{advantage} across the math, knowledge-intensive, and the instruction following instances compared to vanilla instruct model. 
$\varepsilon$ and $\beta$ are hyperparameters controlling the clipping ratio and the weight of the Kullback–Leibler (KL) divergence penalty \cite{schulman2017proximal,shao2024deepseekmath}, respectively. Specifically, $\varepsilon$ determines the permissible range for policy updates, while $\beta$ regulates the magnitude of the KL penalty during training to prevent excessive policy shifts from the reference policy $\pi_{ref}$ (typically the initialization of $\pi_{\theta}$). $D_{KL}\bigl(\pi_\theta \,\|\, \pi_{\text{ref}}\bigr)
= \frac{\pi_{\text{ref}}(y_i \mid x)}{\pi_\theta(y_i \mid x)}
- \log\!\Bigl(\frac{\pi_{\text{ref}}(y_i \mid x)}{\pi_\theta(y_i \mid x)}\Bigr)
- 1 \,$ is the KL divergence approximation term.

%% file: text/5-Experiments.tex
\section{Experiments}
\label{sec: Experiments}

\begin{table*}[t!]
\centering
\small
\setlength{\tabcolsep}{4pt} 
\begin{tabular}{lccccccccccc}
\toprule
\multirow{3}{*}{\textbf{Model}} & \multicolumn{4}{c}{\textbf{Knowledge-Intensive Domain}} & \multicolumn{4}{c}{\textbf{Mathematical Domain}} & \multicolumn{2}{c}{\textbf{Open Domain}} & \multirow{3}{*}{\textbf{AVG}} \\
\cmidrule(lr){2-5} \cmidrule(lr){6-9} \cmidrule(lr){10-11} 
& HotpotQA & 2Wiki & MuSiQ & Bamb & AIME24 & AIME25 & MATH500 & GSM8K & LogiQA & IFEval & \\
& EM & EM & EM & EM & EM & EM & EM & EM & Acc & SAcc & \\
\midrule
Qwen2.5-7B-Instruct & 19.27 & 25.49 & 3.60 & 10.40 & 0.00 & 0.00 & 20.40 & 18.57 & \underline{52.99} & 67.65 & 21.84 \\
\midrule 
Naive RAG & 32.18 & 25.62 & 6.41 & 19.20 & 0.00 & 0.00 & 17.40 & 17.13 & 48.54 & \underline{71.35} & 23.78 \\
Iter-RetGen & 34.65 & 27.81 & 8.23 & 20.00 & 3.33 & 0.00 & 17.18 & 16.83 & 48.69 & \textbf{71.53} & 24.83 \\
IRCoT & 30.52 & 21.29 & 7.16 & 22.40 & 0.00 & 0.00 & 11.80 & 31.46 & 35.79 & 28.84 & 18.93 \\
\midrule
SimpleRL-Zero & 4.42 & 12.03 & 1.37 & 14.40 & 26.67 & \textbf{16.67} & 60.00 & 84.76 & 35.48 & 7.76 & 26.36 \\
Eurus-2-7B-PRIME & 11.52 & 22.53 & 1.82 & 12.00 & 16.67 & 13.33 & \underline{62.00} & \textbf{90.07} & 43.78 & 20.52 & \underline{29.42} \\
\midrule
ToRL & 1.12 & 0.49 & 0.37 & 4.00 & \textbf{33.30} & 10.00 & 58.40 & 81.96 & 39.02 & 13.12 & 24.18 \\
Search-R1 & 35.41 & 31.23 & 15.18 & 40.00 & 13.33 & 3.33 & 36.00 & 56.18 & 47.31 & 14.60 & 29.26 \\
IKEA & 26.75 & 23.51 & 14.23 & 23.20 & 13.33 & 3.33 & 42.40 & 48.14 & 50.23 & 28.65 & 27.38 \\
ReSearch & \underline{42.17} & \textbf{44.79} & \underline{21.27} & \underline{41.60} & 0.00 & 0.00 & 32.00 & 47.54 & 37.94 & 19.22 & 28.65 \\
\midrule
\textbf{AutoTIR} & \textbf{43.15} & \underline{44.47} & \textbf{23.58} & \textbf{43.20} & \textbf{33.33} & \textbf{16.67} & \textbf{62.60} & \underline{88.48} & \textbf{53.56} & 51.02 & \textbf{46.01} \\
\bottomrule
\end{tabular}
\caption{Performance of baselines across 10 benchmarks. The top two results are highlighted in bold and underlined. All baselines utilize Qwen2.5-7B (Base or Instruct) as the backbone model.  Dataset abbreviations include 2Wiki (2WikiMultiHopQA), MuSiQ (MuSiQue), and Bamb (Bamboogle).
}
\label{tab:my_results} 
\end{table*}

\subsection{Experiment Setup}

\noindent \textbf{Training Settings.} 
We train AutoTIR based on Qwen2.5-7B-Instruct \citep{qwen2.5}. During the RL training stage, we strategically curate our dataset to impart diverse tool-use capabilities and maintain language modeling proficiency. Specifically, the training set of MuSiQue \citep{trivedi2022musique} is incorporated to teach the model how to use retrieval tools for knowledge-intensive reasoning. 
To enhance code tool invocation, we leverage training data from ToRL \citep{li2025torl} and Math-DAPO \citep{yu2025dapo}. Furthermore, to prevent tool overuse and preserve core language modeling abilities, we integrate data from Natural Questions (NQ) \citep{kwiatkowski2019natural}, and instruction-following dataset from \citep{lambert2024tulu}. Notably, for NQ data, we follow the methodology of IKEA \citep{huang2025reinforced}, utilizing samples that the base model can correctly answer without tools, which encourages the model to directly answer simpler questions and thus avoid unnecessary tool invocations. Practically, we find simply setting $r_\text{penalty}=-1$ could be acceptable. More details are provided in the Appendix.

\noindent \textbf{Performance Evaluation Metrics.}  
For mathematical reasoning and open-domain QA tasks, we employ the Exact Match (EM) indicator for measuring model performance, where a prediction is considered correct only if it exactly matches the ground truth answer.  
For the multiple-choice LogiQA dataset \citep{liu2021logiqa}, we use standard Accuracy as the evaluation metric.  
For the instruction-following task, we follow the IFEval benchmark and report soft accuracy (SAcc), which measures the proportion of individual constraints satisfied by the model's response for each query.
To further evaluate the effectiveness and efficiency of tool usage during inference, we employ the following two auxiliary metrics:

\noindent\textbf{Tool Selection Accuracy (TS).}  
This metric assesses the model's ability to choose the appropriate tool given a query \citep{qu2025tool}. Let \( T = \{t_1, t_2, \ldots\} \) denote the set of all tool invocation contents across queries.  
For each tool invocation \( t_i \), let \( E(t) \) represent the tool selected by the model, and \( E^*(t) \) be the ground-truth tool.  
The TS is computed as:
\begin{equation}
    \operatorname{TS} = \frac{1}{|T|} \sum_{t \in T} \mathbb{I}(E(t) = E^*(t))
\end{equation}
where \( \mathbb{I} \) is the indicator function that returns 1 if the selection is correct and 0 otherwise.

\noindent\textbf{Tool Productivity (TP).}  
This metric captures how efficiently the model converts tool usage into correct answers \citep{wang2025otc}.  
It is defined as the number of correct predictions per unit of tool usage:
\begin{equation}
    \text{TP} = \frac{\sum_{i=1}^{N} \mathbb{I}\{y_{i} = \hat{y}_{i}\}} 
    {1 + \sum_{i=1}^{N} |t_{i}|}
\end{equation}
where \( \hat{y}_i \) is the predicted answer, \( y_i \) is the ground truth, and \( |t_i| \) denotes the number of tool invocations in the \( i \)-th instance.  
TP balances correctness and cost, discouraging overuse of tools while rewarding effective tool integration.

\noindent \textbf{Benchmarks.} 
To comprehensively evaluate the tool-use capabilities of our model, we conduct experiments on three categories of datasets:
(1)  \textbf{Knowledge-intensive} reasoning benchmarks, including 
HotpotQA \citep{yang2018hotpotqa}, 2WikiMultiHopQA \citep{ho2020constructing}, MuSiQue \citep{trivedi2022musique}, and Bamboogle \citep{press2023measuring}. Specifically, HotpotQA, 2WikiMultiHopQA, and MuSiQue are constructed among wikipedia or wikidata, via different multi-hop mining strategies with crowd-sourcing, while Bamboogle is a manually constructed dataset with 2-hop questions, where all questions are sufficiently difficult to be unanswerable by a popular internet search engine.
(2)  \textbf{Math reasoning} benchmarks,
including AIME2024, AIME2025, MATH500 \citep{hendrycks2measuring}, and GSM8K \citep{cobbe2021training}.
(3)  \textbf{Open domain} reasoning and instruction following benchmarks, including
LogiQA \citep{liu2021logiqa} and IFEval \citep{zhou2023instruction}. These benchmarks are designed to evaluate the model’s core language modeling abilities, such as logical reasoning, instruction following.

\noindent \textbf{Baselines.}
We include three categories of baselines for comparison:
(1) \textbf{Text-only reasoning models trained with reinforcement learning}:
SimpleRL-Zero \citep{zeng2025simplerl} and Eurus-2-7B-PRIME \citep{cui2025process} enhance base LLMs using RL to improve general reasoning performance, particularly on math-related tasks.
(2) \textbf{Code-enhanced models}, which integrate programmatic tools to support symbolic computation and numerical reasoning:
ToRL \citep{li2025torl} incorporates code execution environments to solve math problems more accurately via tool-based reasoning.
(3) \textbf{Retrieval-enhanced models}, which equip LLMs with access to external textual information:
Naive RAG: A basic retrieval-augmented method that concatenates retrieved passages with the input question for answer generation.
Iter-RetGen \citep{shao2023enhancing}: An iterative framework alternating between retrieval and generation steps.
IRCoT \citep{trivedi2023interleaving}: A multi-step method combining retrieval with chain-of-thought reasoning.
Search-R1 \citep{fu2025scaling}, ReSearch \citep{chen2025learning}, and IKEA \citep{huang2025reinforced}: Reinforcement learning-based methods that learn to invoke retrieval tools during reasoning.

\subsection{Main Results}

Table~\ref{tab:my_results} presents the performance of AutoTIR and three categories of baseline methods: text-only reasoning models, code-enhanced models, and retrieval-enhanced models. Evaluations are conducted across the knowledge-intensive, mathematical, and open domain reasoning and instruction following benchmarks. We have the following observations:

\noindent \textbf{The Necessity of Tool Invocation:}
Compared to the vanilla instruction model, tool-integrated models show significantly improved performance upon corresponding domains. Specifically, the code-enhanced method ToRL shows higher accuracy on the math-reasoning domain, retrieval-enhanced methods such as Naive RAG, Iter-RetGen, Search-R1 and IKEA etc., incorporate external knowledge so that better performance on the knowledge-intensive reasoning domain could be obtained. These results highlight the necessity of integrating tools to complement the shortcomings of LLMs.  
Compared to prior baselines, AutoTIR consistently shows better performance on both the knowledge-intensive domain and the math domain. 

\noindent \textbf{Necessity of Balancing Tool Invocation with Instruction Following}
However, the enhanced reasoning ability of tool integration is generally at the cost of instruction-following ability. On open domain instruction following benchmark IFEval, the tool-integrated methods including ToRL, SearchR1, IKEA, and ReSearch show significant performance degradation. The limitations in following user instructions would severely impact the practical application of these tool-integrated reasoning (TIR) models, highlighting the necessity of balancing the instruction following ability and tool invocation. 
However, how to balance the trade-off between language modeling and tool integration, and discriminate which tool is appropriate increase the complexity of this task, as shown in Table~\ref{tab:tool_efficiency} and corresponding analysis.

\noindent \textbf{Effectiveness Analysis of AutoTIR:}
Compared to State-of-the-Art baselines, AutoTIR achieved the highest average score (46.01) across benchmarks from three domains. Compared to baseline tool-integrated reasoning methods that are limited to either the math-domain or the knowledge-intensive domain, AutoTIR preserves the instruction following ability, meanwhile showing better performance on both the math-domain or the knowledge-intensive domain. These results show that, LLMs have the potential to balance tool-integrated reasoning with language modeling and can learn to appropriately decide whether and which tool to invoke through reinforcement learning. 
Note that all benchmarks besides MuSiQue have different sources compared to the training data. These results indicate that AutoTIR can learn to autonomously invoke external tools and generalize to out-of-distribution test sets, with the guidance of the advantage-based reward system.

\begin{table}[t]
\centering
\resizebox{\linewidth}{!}{
\begin{tabular}{lccccc}
\toprule
\multirow{2}{*}{\textbf{Model}} & \textbf{HQA} & \textbf{2Wiki} & \textbf{MATH} & \textbf{GSM8K} & \textbf{IFEval} \\
 & \textbf{EM} & \textbf{EM} & \textbf{Acc} & \textbf{Acc} & \textbf{SAcc} \\
\midrule
\textbf{AutoTIR} & 43.15 & 44.47 & 62.6 & 88.48 & 51.02 \\
- w/o Tools & 22.48 & 25.37 & 56.6 & 83.70 & 34.01 \\
- w/o IF & 40.57 & 44.67 & 63.4 & 85.29 & 13.12 \\
- w/o Penalty & 42.98 & 42.64 & 58.2 & 87.79 & 47.13 \\
- w/ Prior & 42.36 & 43.12 & 57.2 & 86.58 & 44.36 \\
\bottomrule
\end{tabular}
}
\caption{Ablation study on AutoTIR}\label{fig:ablation}
\end{table}

\begin{figure*}[h]
   \begin{center}
   \includegraphics[width=0.85\linewidth]{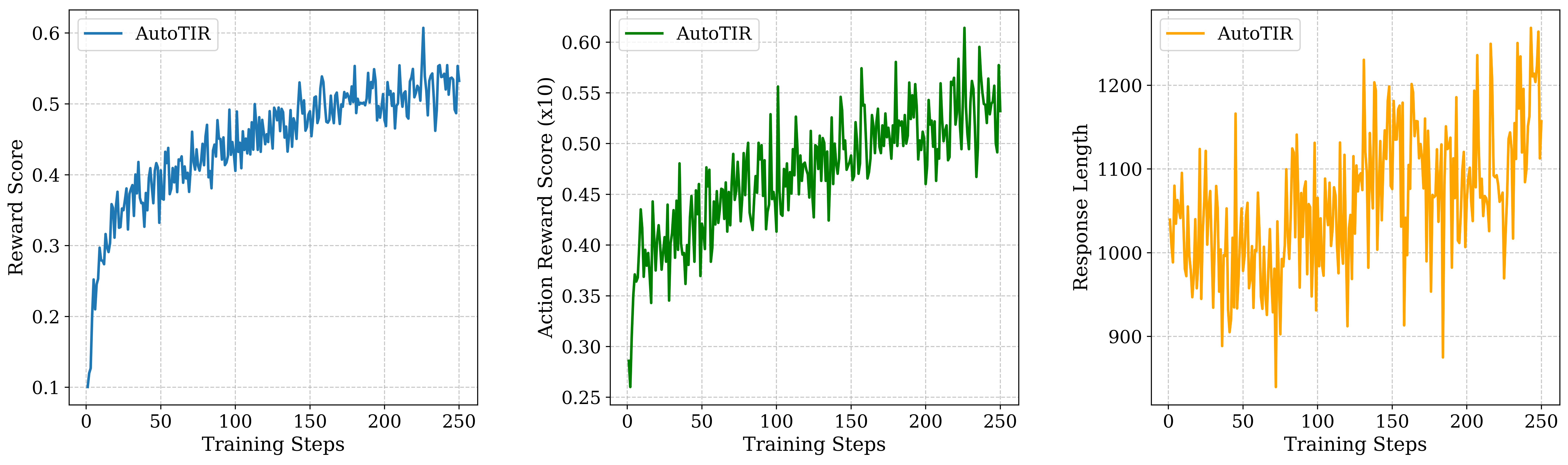}
   \end{center}
   \caption{Avg. reward score and response length during training.
   }
   \label{fig:response_length_reward}
\end{figure*}

\begin{table*}[h]
\centering
\small
\setlength{\tabcolsep}{4pt} 
\begin{tabular}{lcccccccccccccccc}
\toprule
\multirow{2}{*}{Models} & \multicolumn{2}{c}{\textbf{HotpotQA}} & \multicolumn{2}{c}{\textbf{2Wiki}} & \multicolumn{2}{c}{\textbf{MuSiQue}} & \multicolumn{2}{c}{\textbf{Bamboogle}} & \multicolumn{2}{c}{\textbf{LogiQA}} & \multicolumn{2}{c}{\textbf{IFEval}} & \multicolumn{2}{c}{\textbf{AVG}} \\
\cmidrule(lr){2-3} \cmidrule(lr){4-5} \cmidrule(lr){6-7} \cmidrule(lr){8-9} \cmidrule(lr){10-11} \cmidrule(lr){12-13} \cmidrule(lr){14-15}
& TS $\uparrow$ & TP $\uparrow$ & TS $\uparrow$ & TP $\uparrow$ & TS $\uparrow$ & TP $\uparrow$ & TS $\uparrow$ & TP $\uparrow$ & TS $\uparrow$ & TP $\uparrow$ & TS $\uparrow$ & TP $\uparrow$ & TS $\uparrow$ & TP $\uparrow$\\
\midrule
SearchR1 & 99.78 & 16.82 & 99.92 & 12.38 & 99.53 & 6.01 & 99.20 & 19.12 & 71.25 & 44.80 & - & 53.95 & 93.94 & 25.51 \\
IKEA & 93.01 & 18.82 & 94.24 & 14.57 & 99.50 & 7.38 & 91.77 & 18.35 & 45.29 & 48.48 & - & 38.74 & 84.76 & 24.39 \\
ReSearch & 98.07 & 18.04 & 98.76 & 15.99 & 97.92 & 7.48 & 97.41 & 18.89 & 0.00 & 8.22 & - & 15.74 & 78.43  & 14.06 \\
\textbf{AutoTIR} & 92.34 & 20.05 & 91.31 & 17.03 & 95.93 & 8.63 & 94.83 & 19.19 & 97.86 & 52.75 & - & 54.91 & 94.45 & 28.76 \\
\bottomrule
\end{tabular}
\caption{Tool-use efficiency of AutoTIR and baselines on knowledge-intensive reasoning and general domain tasks.}
\label{tab:search_tool}
\end{table*}

\begin{table}[h]
\centering
\resizebox{\linewidth}{!}{
\begin{tabular}{lcccccccc}
\toprule
\multirow{2}{*}{Models} & \multicolumn{2}{c}{AIME25} & \multicolumn{2}{c}{GSM8K} & \multicolumn{2}{c}{IFEval} \\
\cmidrule(lr){2-3} \cmidrule(lr){4-5} \cmidrule(lr){6-7}
& TS & TP & TS & TP & TS & TP \\
\midrule
ToRL & 85.71 & 8.57 & 98.42 & 80.59 & - & 49.23 \\
AutoTIR & 100.00 & 6.67 & 100.00 & 76.15 & - & 54.91 \\
\bottomrule
\end{tabular}
}
\caption{Tool-use efficiency of AutoTIR against ToRL.} 
\label{tab:tool_efficiency}
\end{table}

\subsection{Ablation Analysis}
We conduct an ablation study to assess the contribution of each key component of AutoTIR. In which, w/o Tools refers to prohibiting AutoTIR from utilizing tools via prompting. The prompt is shown in Appendix \ref{app:prompt}. w/o IF refers to removing instruction following data from the RL process; w/o Penalty refers to removing the incorrect tool invocation penalty within the action reward; while w/ Prior refers to forcing the model to utilize tools upon prior set types of instructions, including generation length, key words frequency, and letter frequency. Heuristically, properly utilizing tools, especially code, would directly derive an accurate result. Table \ref{fig:ablation} shows the performance of removing different components of AutoTIR.
From Table~\ref{fig:ablation} we observe:
(1) Impact of Tool Integration: 
Prohibiting tool utilization (w/o Tools) significantly degrades performance across all benchmarks, particularly in knowledge-intensive and mathematical domains. This suggests that without external tools, the performance improvement compared to the vanilla instruction model would be limited, highlighting the necessity of tool integration. 

(2) Role of Instruction-Following Data: 
Excluding instruction-following data during the RL phase (w/o IF) significantly impairs the model's general instruction adherence, as evidenced by a substantial drop on IFEval (from 51.02 to 13.12 SAcc). 
While performance on knowledge-intensive and mathematical tasks remains robust, this specific degradation underscores the critical role of instruction-following data in maintaining and enhancing the model's core language modeling and adherence to instructions during RL training. Moreover, the improvement in instruction following ability does not impair the reasoning ability of AutoTIR, showing the immense potential of LLM in balancing TIR and language modeling. 

(3) Effects of the Tool Action Penalty: 
Removing the penalty term from the reward function (w/o Penalty) leads to performance degradation, particularly on knowledge-intensive tasks like 2WikiMultiHopQA and on instruction following (e.g., IFEval). This shows that incorrect or unnecessary tool invocation are harmful for model performance, and thus the necessity of the action reward to guide the model for making more reasonable and efficient tool selection actions.

(4) Impact of Autonomous Tool Invocation Exploration: 
For open domain instructions lack of a clear standard to decide whether a tool is necessary, we adopt a data-driven paradigm: In the advantage-based reward system, on such instructions, the LLM is encouraged to make a free exploration about whether external tool(s) is necessary without additional action reward or penalty. The performance of w/ prior shows the necessity of such a data-driven paradigm: after forcing the model to utilize tools on 4 types of open domain instructions that the code tool is previously deemed to be helpful, surprisingly, the performance on instruction following benchmark IFEval, as well as math benchmarks GSM8K and MATH500 in turn decreases. In other words, the instruction following ability is impaired. This suggests that, due to the complexity of the distribution of open-domain instructions, and the difference between model knowledge, model reasoning mechanism with that of humans, the human prior may not always be helpful in improving the model's performance. 
Thus, on such questions, free exploration of tool invocation would be necessary, suggesting the reasonability of our advantage-based reward system.

\subsection{Scalability of Performance}

Figure~\ref{fig:response_length_reward} shows the values of reward on the development set during the training process. With more training steps, the action reward as well as the output reward continuously increase, indicating that the model is gradually learning the tool invocation strategy and henceforth derives a correct result, in a scalable manner. Our system shows the potential to obtain a model with better performance with more available training instances. Moreover, with the training process, the simultaneous increase of response length and output reward suggests that the model is acquiring a more complex reasoning pattern, so as to adapt to ``hard'' reasoning problems. Detailed reward curve on different dev sets are provided in the Appendix.

\subsection{Tool-Utilization Efficiency Analysis}
To assess AutoTIR's efficiency in tool invocation during stepwise reasoning, we compare its performance against baseline tool-integrated reasoning methods on knowledge-intensive, mathematical, and general reasoning datasets. 
As shown in Table~\ref{tab:search_tool} and \ref{tab:tool_efficiency}, we analyze two key tool-usage metrics: Tool Selection (TS), which measures the accuracy of choosing the appropriate tool, and Tool Productivity (TP), which evaluates how effectively the chosen tool aids in reaching a correct final answer. On open domain instructions, TS is not calculated, as there is no clear standard for judging if tool(s) should be invoked.

(1) The tool selection accuracy (TS) and tool productivity (TP) of baseline TIR methods are task-specific and lack of generalizability. 
On the knowledge-intensive domain, retrieval-based baselines like SearchR1 and ReSearch show higher TS scores. However, their TS performance significantly drops on open domain datasets such as LogiQA and IFEval. Similarly, the code-enhanced TIR method ToRL fails to appropriately and efficiently invoke tools for deriving a correct answer on instruction following benchmarks. This show the challenge of learning a generalizable tool invocation strategy to adapt across different domains.

(2) AutoTIR can learn a generalizable tool invocation strategy to appropriately utilize tools across different domains. On both the knowledge-intensive and the math domain, AutoTIR show comparable or higher tool selection accuracy and tool productivity compared to SoTA baselines. This shows that, AutoTIR learns a generalizable strategy to autonomously determine whether and which tools to use during reasoning across different domains. 
This learned strategic decision-making proves particularly beneficial in complex and diverse reasoning tasks, allowing AutoTIR to judiciously leverage tools only when truly necessary for the task, thereby optimizing both performance and efficiency.

\begin{figure}[h]
   \begin{center}
   \includegraphics[width=0.95\linewidth]{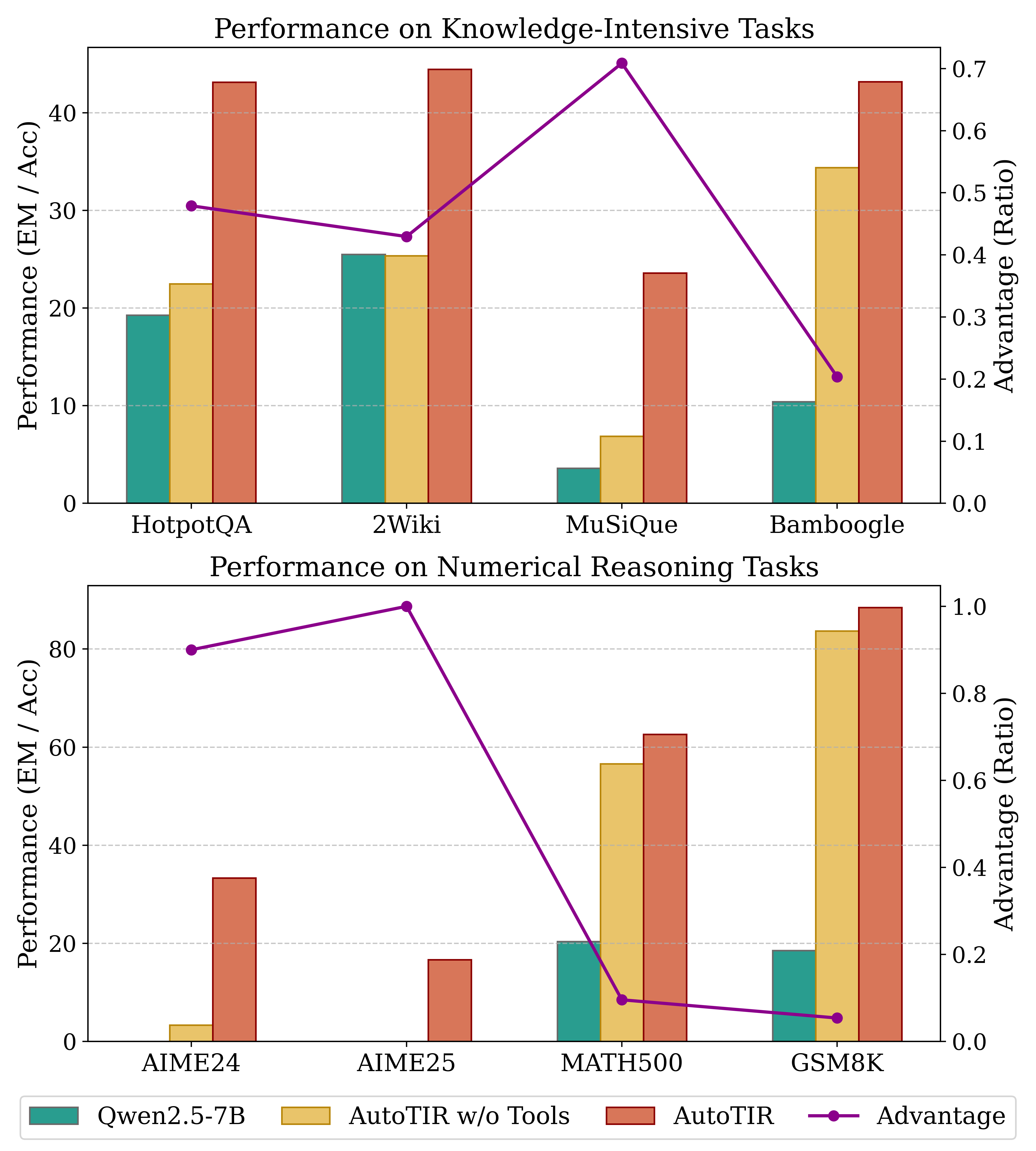}
   \end{center}
   \caption{Model Performance and Tool Advantage Across Reasoning Task Types.
   }
   \label{fig:advantage}
\end{figure}

\subsection{Tool Advantage Analysis}
As shown in Figure \ref{fig:advantage}, we compare the performance advantage gained by tool-integration in the AutoTIR model, relative to its tool-agnostic counterpart. This advantage is measured using an advantage ratio, calculated by dividing the performance improvement from the tools by the overall performance of AutoTIR. To realize the ``tool-agnostic" counterpart, denoted as ``AutoTIR w/o Tools," 
Our analysis reveals several key insights as following:

\noindent \textbf{Consistent Performance Gains}: AutoTIR with integrated tools consistently boosts reasoning performance across both knowledge-intensive and mathematical reasoning tasks. This highlights the general effectiveness of tool utilization.
\noindent \textbf{Reinforcement Learning Efficacy}: Even without tool assistance, AutoTIR, which degrades to an R1-like RL model \citep{guo2025deepseek}, still outperforms the baseline model Qwen2.5-7B-Instruct. This strongly suggests that RL training inherently enhances the model's reasoning capabilities.
\noindent \textbf{Varying Tool Impact by Task Difficulty}:
The performance advantage derived from tool usage varies significantly with task difficulty. For simpler numerical reasoning tasks like GSM8K (elementary school level) and MATH500 (middle school level), the improvements from tools are relatively modest. In contrast, for more challenging competitive mathematical problems such as AIME, tool integration provides a substantial performance boost. This indicates that tools are particularly beneficial for complex problem-solving.

Our approach aims to optimize overall performance by strategically invoking tools where they offer the most significant advantage. Crucially, AutoTIR consistently showed stronger results in this domain compared to other RL baselines, suggesting a more effective balance between specialized reasoning and general instruction following.

%% file: text/7-Conclusion.tex
\section{Conclusion}

Current Tool-Integrated Reasoning (TIR) approaches often rely on rigid tool-use patterns, limiting flexibility and undermining core language abilities. We propose \textbf{AutoTIR}, a reinforcement learning framework that enables LLMs to autonomously determine \emph{whether} and \emph{which} tools to invoke based on task context.
AutoTIR introduces a hybrid reward mechanism that jointly optimizes for answer correctness, structured output adherence, and penalization of incorrect tool usage. This design fosters precise reasoning while reducing unnecessary tool interactions.
Extensive evaluations across knowledge-intensive, mathematical, and general domain benchmarks show that AutoTIR achieves superior overall performance, significantly outperforming baselines and generalizes well across tasks. Its learned tool-use policy dynamically adapts to varying task demands, achieving more efficient and context-aware tool integration.
By shifting from static invocation patterns to adaptive, learned strategies, AutoTIR offers a scalable and generalizable foundation for future tool-augmented LLMs.

%% file: text/8-Appendix.tex
\section{Appendix }

\subsection{System Prompt Template for AutoTIR}
\label{app:prompt}

\begin{figure*}[ht]
   \begin{center}
   \includegraphics[width=0.95\linewidth]{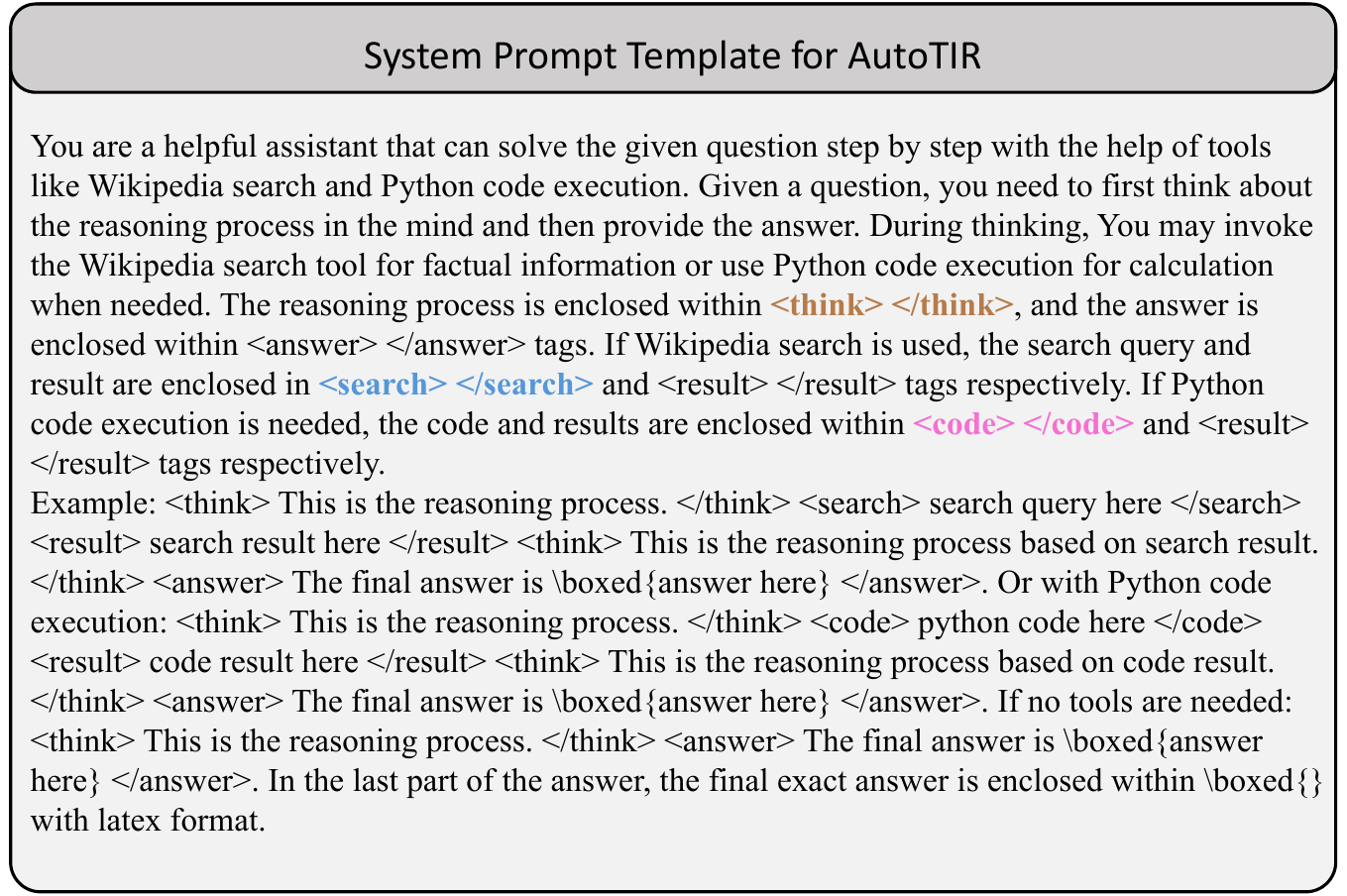}
   \end{center}
   \caption{System prompt template for training and inference from AutoTIR.}
   \label{fig:prompt_tool}
\end{figure*}

\begin{figure*}[ht]
   \begin{center}
   \includegraphics[width=0.95\linewidth]{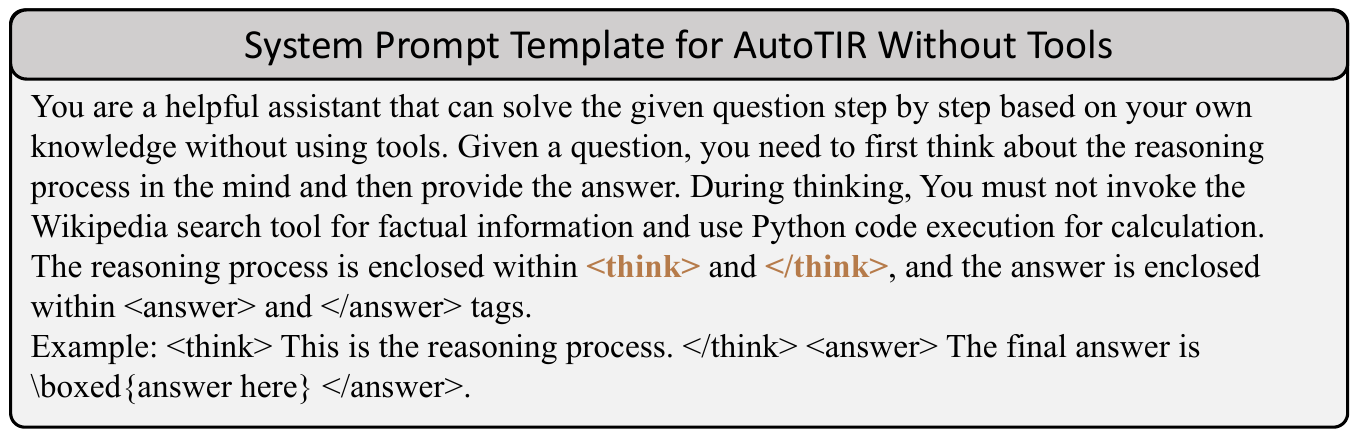}
   \end{center}
   \caption{System prompt template for text-based inference from AutoTIR.}
   \label{fig:prompt_without_tool}
\end{figure*}

Our rollout process relies on special control tags (e.g., \texttt{</search>} and \texttt{</code>}) to trigger tool invocation. To ensure proper policy execution, the LLM must strictly adhere to the predefined output format.
In addition, we designed two distinct prompt templates for AutoTIR (Figure \ref{fig:prompt_tool} and \ref{fig:prompt_without_tool}), enabling dynamic switching between tool-assisted reasoning and standalone reasoning:
Tool-assisted mode leverages external tools for complex tasks requiring retrieval or code execution.
Standalone mode bypasses tool use for simpler tasks (e.g., GSM8K), prioritizing inference speed.
Inspired by DeepSeek-R1, these templates ensure smooth policy adaptation during RL training. This dual-mode design optimizes efficiency: while tool integration improves reasoning in complex cases, avoiding unnecessary tool calls for simple problems significantly reduces latency, critical for tasks where tool overhead outweighs benefits.

\begin{figure*}[t]
   \begin{center}
   \includegraphics[width=1\linewidth]{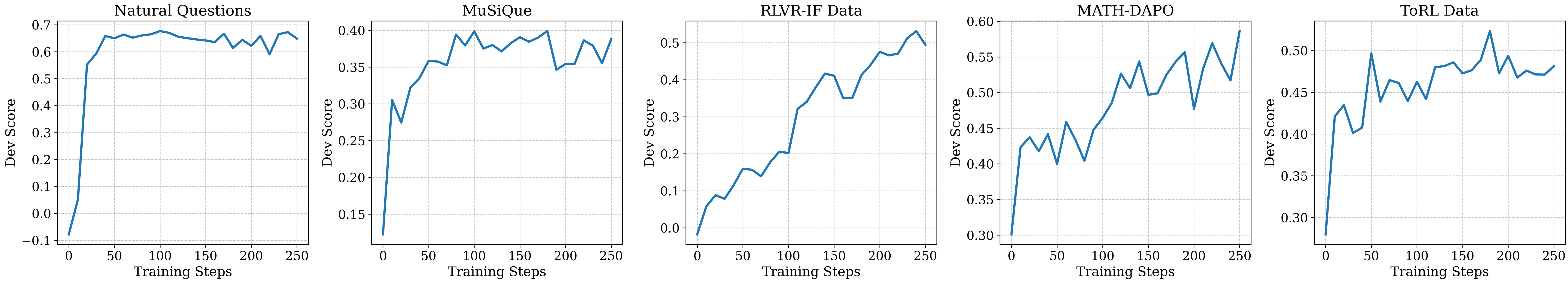}
   \end{center}
   \caption{Performance metrics across different datasets during training stage.
   }
   \label{fig:dev_performance}
\end{figure*}

\subsection{Test-Time Scaling Analysis}
\begin{figure*}[t]
   \begin{center}
   \includegraphics[width=1\linewidth]{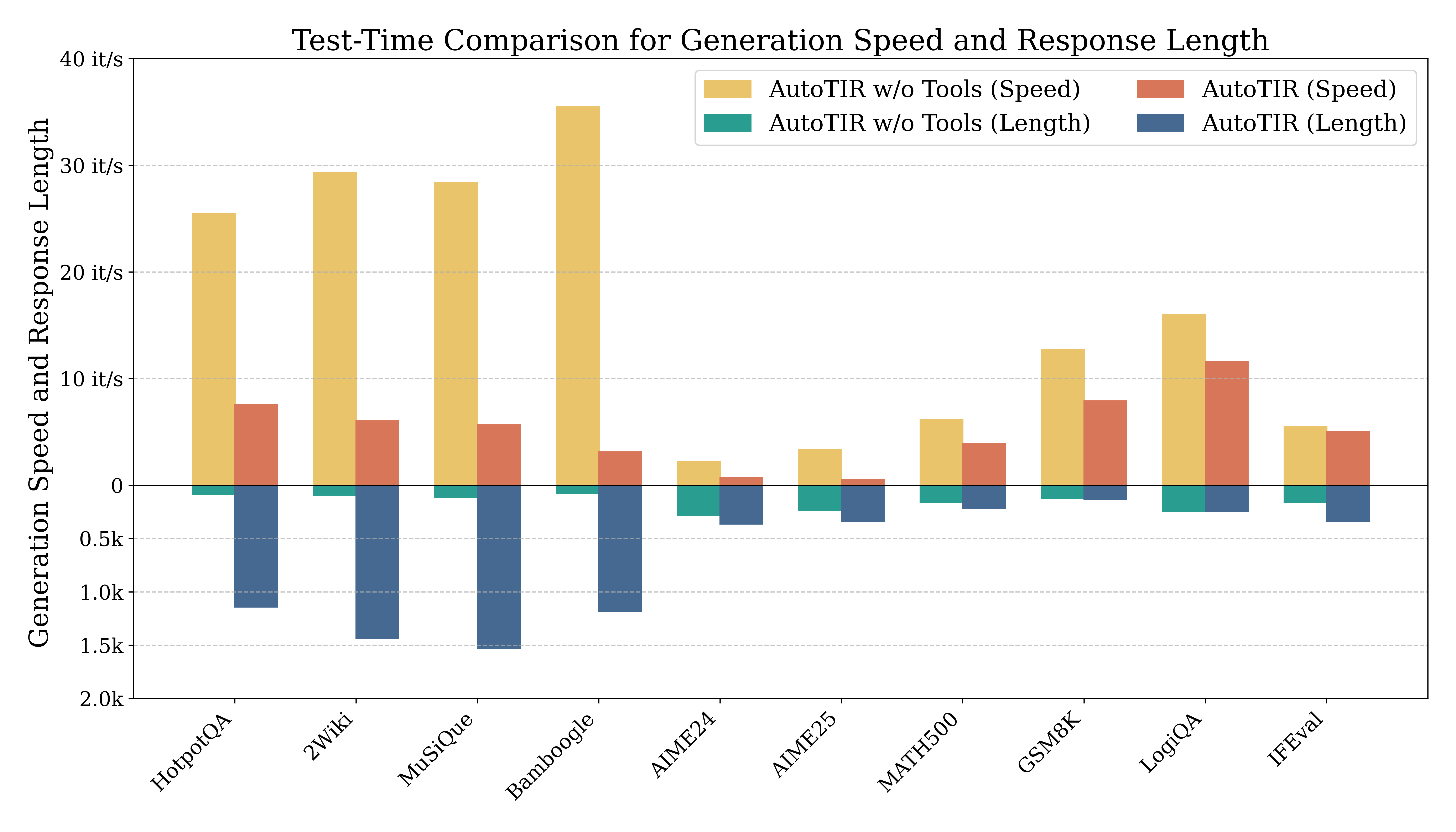}
   \end{center}
   \caption{Test-Time comparison.
   }
   \label{fig:test-time}
\end{figure*}

Building on our performance analysis, we examine the computational efficiency of AutoTIR at test time. While AutoTIR can selectively employ tools based on problem categories, its current approach shows inefficiency in invoking tools based on single-sample difficulty. This can lead to unnecessary computational overhead.

As depicted in Figure \ref{fig:test-time}, this inefficiency is evident. For simpler tasks like GSM8K and MATH500, the `AutoTIR w/o Tools' configuration achieves significantly faster inference speeds. This highlights the direct computational cost of tool invocation where performance gains are minimal. Furthermore, `AutoTIR w/ Tools' consistently generates substantially longer responses across all tasks (e.g., 1.5k tokens for MuSiQue with tools vs. 0.1k without), contributing to higher generation costs and slower inference times.

These findings underscore the value of an adaptive "tool-integrated reasoning switch." Such a mechanism would enable AutoTIR to bypass tool calls for straightforward problems, optimizing compute resources and achieving faster inference without compromising performance, thus enhancing practical deployment efficiency.

\subsection{Scaling Analysis on the RL Training Curve}
This section examines key metrics during the training of AutoTIR. Specifically, Figure \ref{fig:response_length_reward} illustrates the response length and training reward progression during training.

\noindent \textbf{Response Length.} We define response length as the total number of tokens in the model's output, excluding retrieval results. This metric can be interpreted as the test-time cost of reasoning. Figure \ref{fig:response_length_reward} clearly shows a general increase in response length throughout the training process. This suggests AutoTIR progressively acquires long CoT reasoning capabilities during this phase.

\noindent \textbf{Training Reward.} Furthermore, the format reward improves with increasing training steps, indicating that AutoTIR begins generating responses adhering to predefined formats. Through the overall improvement in reward scores, the system ultimately learns an action strategy across different task types, determining whether to use tools and selecting which specific tool to employ for assisted reasoning.

\noindent \textbf{Training Performance.} The training of our AutoTIR model involves a reinforcement learning (RL) process, carefully designed to enhance its tool-use capabilities and instruction-following abilities across diverse reasoning tasks. During training, the performance of the model is monitored on several validation benchmarks, as depicted in the Figures \ref{fig:dev_performance}.
Specifically, we observe the training progress across five distinct datasets:
Each of these datasets represents a unique challenge, ranging from knowledge-intensive question answering (e.g., NQ, MuSiQue) and mathematical reasoning (e.g., MATH-DAPO \citep{yu2025dapo}, ToRL data \citep{li2025torl}) to general instruction following (e.g., RLVR-IF data \citep{lambert2024tulu}).
As the training steps progress up to 250, AutoTIR consistently demonstrates a positive performance trend across nearly all benchmarks. We generally observe a rapid initial improvement in dev scores, indicating the model quickly learns to leverage tools and follow instructions more effectively. This initial steep ascent is often followed by a gradual stabilization or continued, albeit slower, improvement. This trend suggests that the RL optimization successfully guides the model towards higher efficacy in utilizing external tools and adhering to task-specific instructions. The consistent upward trajectory across various domains underscores AutoTIR's robust and generalizable learning capabilities during the RL fine-tuning phase.

\subsection{Implementation Details}
Our experiments are conducted using 8 NVIDIA A800-80G GPUs. We employ Qwen2-7B-instruct as the base model, trained with the GRPO reinforcement learning algorithm within the verl framework \cite{sheng2025hybridflow}. For the search tool, we utilize e5-base-v2 \cite{wang2022text} against a Wikipedia 2018 corpus. Baseline results are reproduced using FlashRAG \cite{jin2025flashrag}. Key parameter settings are detailed in Table \ref{tab:AutoTIR_implementation}.

\begin{table}[htbp]
\centering
\begin{tabular}{lc}
\toprule
\textbf{Parameter} & \textbf{Value} \\
\midrule
Learning Rate & 1e-6 \\
Train Batch Size & 256 \\
Number of Training Epochs & 2 \\
Number of Rollout & 5 \\
Rollout Temperature & 1.0 \\
KL Loss Coefficient & 0.001 \\
Clip Ratio & 0.2 \\
\bottomrule
\end{tabular}
\caption{Implementation details of AutoTIR.}
\label{tab:AutoTIR_implementation}
\end{table}